# Head Gesture Recognition using Optical Flow based Classification with Reinforcement of GMM based Background Subtraction

Parimita Saikia
Department of Electronics and Communication Engineering
DBCET, Assam Don Bosco University

Karen Das
Department of Electronics and Communication Engineering
DBCET, Assam Don Bosco University

## ABSTRACT
This paper describes a technique of real time head gesture recognition system. The method includes Gaussian mixture model (GMM) accompanied by optical flow algorithm which provided us the required information regarding head movement. The proposed model can be implemented in various control system. We are also presenting the result and implementation of both mentioned method.

## 1. INTRODUCTION
A primary goal of gesture recognition is to implement mathematical algorithms so that a computer can identify gestures in a efficient, powerful, and flexible way. Human gestures provide one of the most important means for non-verbal interaction among peoples. Automatic gesture recognition, particularly computer vision based techniques do not require the user to wear extra sensors, clothing or equipment for the recognition system.

There are a lot of devices which are applied to sense body position and orientation, facial expression and other aspects of human behavior or state which can be used to determine the communication between the human and the environment. Combinations of human body and sensing devices can produce a wide range of interfacing techniques. To support gesture recognition, human body movement must be tracked and interpreted in order to recognize the meaningful gestures. Many methods have been developed for both hand gesture and body gesture recognition [22][9][4]. These recognition tasks are challenging because human body is highly articulated. In our work of gesture recognition we used Gaussian Mixture Model (GMM)[27] for background subtraction, and background subtracted image is used for further processing. For the determination of head movement we implemented optical flow method [12]. There are a lot of optical flow methods, among which we tried the Horn-Schunck optical flow algorithm [20].

The main goal of the project is to design a real time head movement detection and gesture recognition system. To support gesture recognition, human position and movement must be tracked and interpreted in order to recognize the meaningful gestures [23]. In this paper, we present a gesture recognition system that takes input from a simple camera and necessary processing steps is done to recognize the gestures from a live video.

## 2. BACKGROUND
There has been a tremendous effort in research toward devices and techniques to improve the interaction of human with computers [24]. Gesture recognition is among the most promising attempt in this direction. Gestures are expressive, meaningful body motions involving physical movements of the head, fingers, hands, arms, etc., with the intent of conveying meaningful information or interacting with the environment.

First attempts to solve this problem resulted in mechanical devices that directly measure head, hand and arm joint angles and spatial position. This group is best represented by the so-called glove-based devices. Glove-based gestural interfaces require wearing a cumbersome device, and generally carrying a load of cables. This limits the ease and naturalness with which the user can interact. The problem of having to wear gloves can be overcome by using video-based interaction techniques. In recent years, however, there has been an interest in incorporating the dynamic characteristics of gestures, which carry as many information as postures.

One of the first gesture recognition systems was presented by Maggioni and Rottger (1999) at Siemens. This development uses a video projector which displays a user interface onto any surface and a video camera to capture the hand of a user. By moving the hand, the user can move objects on the projected desktop. This interface uses dynamic hand movements to control the system [25]. A static gesture based system was presented by BMW and is described in Akyol et al. (1999). The system is used to control the infotainment inside a vehicle. Currently, there are several available techniques that are applicable for hand gesture recognition, which are either based on auxiliary devices or computer vision. In the present research effort, we considered these aspects by taking it as a reference to a smart interaction environment of virtual object manipulation and control. Here we executed the head movement detection system which later on translated into command for further. Gestures, head movements, facial expressions – all of these can be valuable tools when skillfully employed. The more communication methods we employ, the more effectively we will communicate..

A lot of work has been continuing in the field of gesture recognition system. Though the demand in the field if gesture recognition is increasing continuously some notable works are included here. Among those - Susmita Mitra provided a gesture recognition survey and tools for using gesture recognition, such as HMM, particle filtering and condensation algorithm, PSM approach. A hybridization of HMMs and FSMs is a potential study in order to increase the reliability and accuracy of gesture recognition .HMMs are computationally expensive and require large amount of training data [10]. Miss Shweta K yewale & Mr Pankaj Bharnes provide a ANN approach for recognition system. At present, artificial neural networks are emerging as the technology of choice for many applications, such as pattern recognition, gesture recognition, prediction, system identification, and control. According to them ANN provides good & powerful solution for gesture recognition in MATLAB [9] .Bo Peng & his group tried full body gesture recognition from video. They used multi linear analyzer on the image of each captured video frame in order to obtain view independent feature vectors for the static poses. They use Hidden Markov Model (HMM) to perform gesture recognition. Here a video based full body gesture recognition system is presented by applying view-invariant pose coefficient vectors as feature vectors. This system is independent of the body orientation of the subject performing the gesture with respect to the camera system [20].Qing Chen Nicolas D Georganes, Emil M Petnin made a approach of real time vision based hand gesture recognition using Haar like features. Each Haar like feature consist of two or three connected black and white rectangles .It provided accuracy for more than 50%. They proposed a two level approach to recognize hand gestures in real-time with a single web camera as the input device. The low level of the approach is focused on the posture recognition with Haar-like features and the Ada Boost learning algorithm. The Haar-like features can effectively describe the hand posture pattern with the computation of "integral image" [19].Prateen Chakraborty, Prashant Sarangi & his group made comparative study as Hand gesture recognition system .They provide the method such as subtraction , PCA, Gradient method, rotation invariant method to gesture recognition[1].

.Sujitha Martin, Cuong Tran, Ashish Tawari, Jade Kwan and Mohan Trivedi using optical flow for the detection of head movement. They are using this detection technique for automotive environment [3]. Kazumoto TANAKA tried to do Gesture Recognition with a Focus on Important Actions by Using a Path Searching Method in Weighted Graph. They are using eigenspace method for feature vector sequence generation [2]. H. Sawada, S. Hata, S. Hashimoto in their paper Gesture recognition for human-friendly interface in designer – consumer cooperate design system, describe optical flow based gesture recognition system able to determine hand motion. They use skin color detection to extract the area of a hand, then they extract the contour of the hand area and determine the tips of the fingers. The fingertips are detected because of their sharp curvature. The 3D position of the hand is finally obtained by applying a stereo matching technique to each of the fingertips. They use an optical flow technique applied to the fingertip areas for detecting the 3D motion of the hand. The results are applied to a gesture driven design system, where customer and designer can interactively exchange their opinion for the prototyping. Commands such as pointing, movement, rotation and zooming are recognized by the vision system. Berthold K.P. Horn and Brian G. Shunck are trying to implement a method for finding the optical flow pattern is presented which assumes that the apparent velocity of the brightness pattern varies smoothly almost everywhere in the image. An iterative implementation is shown which successfully computes the optical flow for a number of synthetic image sequences [12]. Rafael A. B. de Queiroz , Gilson A. Giraldi, Pablo J. Blanco, Raúl A. Feijóo also tried to do determine optical flow by using Horn-Schunk algorithm. The tests indicated that the modified Horn and Schunck's algorithm has convergence rate a little superior to the original model. Their experiments also provided different experimental comparison between the algorithms [20].

## 3. BACKGROUND THEORY

In the head movement detection and corresponding gesture recognition system we first tried to concentrate our interest on the main object by subtracting the background and then applied proper classifier to determine the movement.

### 3.1 Gaussian Mixture Model [26][27]

A Gaussian Mixture Model (GMM) is a parametric probability density function represented as a weighted sum of Gaussian component densities. GMMs are commonly used as a parametric model of the probability distribution of continuous measurements or features in a biometric system, All GMM parameters are estimated from training data using the iterative Expectation Maximization (EM) algorithm or Maximum A Posteriori (MAP) estimation from a well-trained prior model.

A Gaussian mixture model is a weighted sum of M component Gaussian densities as given by the equation,

$$p(X|\lambda = \sum_{i=1}^{M} w_i g(X|\mu_i, \Sigma_i))g \qquad (1)$$

where x is a D-dimensional continuous-valued data vector (i.e. measurement or features), $w_i$, i=1, . . . ,M, are the mixture weights, and $g(x|\mu_i, \Sigma_i)$, i = 1, . . . ,M, are the component Gaussian densities. Each component density is a D-variate Gaussian function of the form,

$$g(X|\mu_i, \Sigma_i) = \frac{1}{(2\pi)^{\frac{D}{2}}|\Sigma_i|^{1/2}} \exp\left\{-\frac{1}{2}(X-\mu_i)'\Sigma_i^{-1}(X-\mu_i)\right\} \quad (2)$$

With mean vector $\mu_i$ and covariance matrix $\Sigma_i$. The mixture weights satisfy the constraint that $\sum_{i=1}^{M} w_i = 1$.

#### 3.1.1 Implementation

A mixture of K Gaussians is used to model the time series of values observed at a particular pixel. The probability of occurrence of the current pixel value is given by

$$P(Z) = \sum_{i=1}^{k} w_i N(\mu_t, \Sigma_t, Z) \qquad (3)$$

Where $N$ is the Gaussian probability density function, whose mean vector is $\mu$ and covariance is $\Sigma$.

And $w_i$ is the weight of the $i^{th}$ Gaussian such that $\Sigma w_i = 1$. The covariance matrix is assumed to be of the form $\Sigma = \sigma^2 I$ for computational reasons.

#### 3.1.2 Parameter Updates

The new pixel value $Z_t$ is checked against each Gaussian. A Gaussian is labeled as matched if

$$\|Z - \mu_h\| < d\sigma_h \qquad (4)$$

Then its parameters may be updated as follows:

$$w_{i,t} = (1 - \alpha) * w_{i,t-1} + \alpha * M_{i,t} \qquad (5)$$

$$\mu_t = (1 - \rho) * \mu_{t-1} + \rho * Z_t \qquad (6)$$

$$\sigma_t^2 = (1 - \rho) * \sigma_{t-1}^2 + \rho * (Z_t - \mu_t)^T * (Z_t - \mu_t) \qquad (7)$$

$$\rho = \alpha * N(\mu_{t-1}, \Sigma_{t-1}, Z_t) \qquad (8)$$

Where α is the learning rate for the weights.

If a Gaussian is labeled as unmatched only its weight is decreased as

$$w_{i,t} = (1 - \alpha) * w_{i,t-1} \quad (9)$$

If none of the Gaussians match, the one with the lowest weight is replaced with $Z_t$ as mean and a high initial standard deviation.

The rank of a Gaussian is defined as w/σ. This value gets higher if the distribution has low standard deviation and it has matched many times. When the Gaussians are sorted in a list by decreasing value of rank, the first is more likely to be background. The first B Gaussians representing the background

$$B = \arg min_b (\sum_{k=1}^{b} w_i > T) \quad (10)$$

The Gaussian mixture model (GMM) is adaptive; it can incorporate slow illumination changes and the removal and addition of objects into the background. The higher the value of T in (10), the higher is the probability of a multi-modal background.

### 3.1.3 Shadow Removal
Shadows which are detected as foreground can cause several problems when extracting the objects, two examples are object shape distortion and several objects merging together. It is especially crucial and these problems should be avoided.

In most traditional methods the image is first transformed to a color space that segregates the chromaticity information from the intensity. In the HSV color space, the hypothesis is that a shadowed pixel value's value and saturation will decrease while the hue remains relatively constant

However, the color model of Horprasert et al.[6][27] gave the best results in the test conducted. It doesn't require any complicated conversion formulae like in the case of HSV. Also the simple choice of parameters is a distinct advantage.

In this model each pixel value in the RGB color space is assumed to lie on a chromaticity line, which connects the pixel value and the origin. The authors specify a way to calculate the deviation of a foreground pixel value from the background value. The foreground value is compared with the means of each of the B background Gaussians in equation (10).

The brightness distortion (BD) and chromaticity distortion (CD) are defined as :

$$BD = \frac{OP}{OB} \ \& \ CD = PF$$

Shadow points (sp) can now be ascertained as:

sp = 1    for α≤ $BD ≤ β \wedge CD ≤ \tau_C$

   =0    otherwise

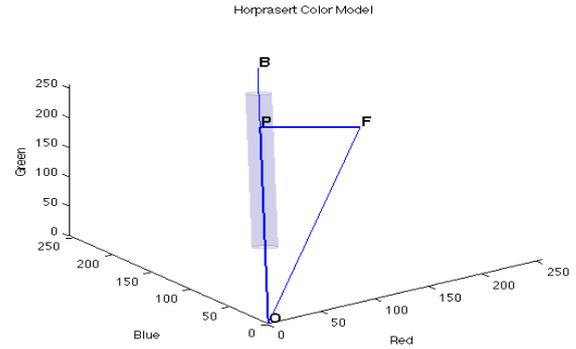

**Figure 1: Brightness and chromaticity distortion in the RGB color space.**

In Figure 1 OB is the background RGB vector and OF is the foreground RGB vector. FP is the perpendicular dropped from F onto OB. The shaded cylinder is the locus of all shadow color values.

## 3.2 Optical flow
The computation of optical flow between two images can be obtained by optical flow constraint equation:

$$I_X u + I_Y v + I_t = 0 \quad (11)$$

In this equation, the following values are represented:

$I_x$, $I_y$ and, $I_t$ are the spatiotemporal image brightness derivatives; *u* is the horizontal optical flow; *v* is the vertical optical flow.

Because this equation is under constrained, there are several methods to solve for *u* and *v*- Horn-Schunck Method and Lucas-Kanade Method.

### 3.2.1 Horn-Schunck Method [12]
We derived an equation that relates the change in image brightness at a point to the motion of the brightness pattern. Let the image brightness at the point (**x,** y) in the image plane denoted by **E(x,** y, t). Now we considered what happens when the pattern moves. The brightness of a particular point in the pattern is constant, so that

$$\frac{dE}{dt} = 0$$

Using the chain rule for differentiation we see that,

$$\frac{dE}{dt} \cdot \frac{\delta E}{\delta x} + \frac{dE}{dt} \cdot \frac{\delta E}{\delta Y} + \frac{\delta E}{\delta t} = 0 \quad (12)$$

If we let,

$$\frac{dx}{dt} = u \quad \text{and} \quad \frac{dy}{dt} = v$$

Then it is easy to see that we have a single linear equation in the two unknowns u and *v*,

$$E_x u + E_y v + E_t = 0 \quad (13)$$

Where we have also introduced the additional abbreviations $E_x$, $E_y$, and $E_t$, for the partial derivatives of image brightness with respect to x, y and *t* respectively. The constraint on the local flow velocity expressed by this equation,

$$(u,v)(E_x, E_y) = -E_t \quad (14)$$

Thus the component of the movement in the direction of the brightness Gradient ($E_x$, $E_y$) equals

$$\frac{E_t}{-E_t\sqrt{(E_x+E_y)}} \tag{15}$$

### 3.2.1.1 Estimating the Partial Derivatives

We estimated the derivatives of brightness from the discrete set of image brightness. It is important that the estimates of $E_x$, $E_y$ and $E_t$ be consistent. That is, they should all refer to the same point in the image at the same time. While there are many formulas for approximate differentiation we used a set which gives us an estimate of $E_x$, $E_y$, $E_t$, at a point in the center of a cube formed by eight measurements. Each of the estimates is the average of four first differences taken over adjacent measurements,

$$E_X \approx \tfrac{1}{4} \{E_{i,j+1,k} - E_{i,j,k} + E_{i+1,j+1,k} - E_{i+1,j,k} + E_{i,j+1,k+1} - E_{i,j,k+1} + E_{i+1,j+1,k+1} - E_{i+1,j,k+1}\} \tag{16}$$

$$E_y \approx \tfrac{1}{4} \{E_{i+1,j,k} - E_{i,j,k} + E_{i+1,j+1,k} - E_{i,j+1,k} + E_{i+1,j,k+1} - E_{i,j,k+1} + E_{i+1,j+1,k+1} - E_{i,j+1,k+1}\} \tag{17}$$

$$E_t \approx \tfrac{1}{4} \{E_{i,j,k+1} - E_{i,j,k} + E_{i+1,j,k+1} - E_{i+1,j,k} + E_{i,j+1,k+1} - E_{i,j+1,k} + E_{i+1,j+1,k+1} - E_{i+1,j+1,k}\} \tag{18}$$

Here the unit of length is the grid spacing interval in each image frame and the unit of time is the image frame sampling period. We avoid estimation formulae with larger support, since these typically are equivalent to formulae of small support applied to smoothed images.

### 3.2.1.2 Estimating the Laplacian of the Flow Velocities

We approximated the Laplacians of $u$ and $v$. One convenient approximation takes the following form,

$$\upsilon \nabla^2 \approx k(\bar{\upsilon}_{i,j,k} - \upsilon_{i,j,k})$$

$$u \nabla^2 \approx k(\bar{u}_{i,j,k} - u_{i,j,k})$$

Where the local averages $\bar{u}$ and $v$ are defined as follows

$$\bar{u}_{i,j,k} = \tfrac{1}{6}\{u_{i-1,j,k} + u_{i,j+1,k} + u_{i+1,j,k} + u_{i,j-1,k}\} + \tfrac{1}{12}\{u_{i-1,j-1,k} + u_{i-1,j+1,k} + u_{i+1,j+1,k} + u_{i+1,j-1,k}\} \tag{19}$$

$$\bar{\upsilon}_{i,j,k} = \tfrac{1}{6}\{\upsilon_{i-1,j,k} + \upsilon_{i,j+1,k} + \upsilon_{i+1,j,k} + \upsilon_{i,j-1,k}\} + \tfrac{1}{12}\{\upsilon_{i-1,j-1,k} + \upsilon_{i-1,j+1,k} + \upsilon_{i+1,j+1,k} + \upsilon_{i+1,j-1,k}\} \tag{20}$$

The proportionality factor $K$ equals 3 if the average is computed as shown and we again assume that the unit of length equals the grid spacing interval.

### 3.2.1.3 Minimization

The problem then is to minimize the sum of the errors in the equation for the rate of change of image brightness,

$$E_x u + E_y v + E_t = \mathcal{E}_b$$

and the measure of the departure from smoothness in the velocity flow,

$$\mathcal{E}_c^2 = \left(\tfrac{\partial u}{\partial x}\right)^2 + \left(\tfrac{\partial u}{\partial y}\right)^2 + \left(\tfrac{\partial v}{\partial x}\right)^2 + \left(\tfrac{\partial v}{\partial y}\right)^2 \tag{21}$$

We determined the relative weight of these two factors. In practice the image brightness measurements will be corrupted by quantization error and noise so that we cannot expect $\mathcal{E}_b$ to be identically zero. This quantity will tend to have an error magnitude that is proportional to the noise in the measurement. This fact guides us in choosing a suitable weighting factor, denoted by $\mathcal{E}^2$.

Let the total error to be minimized by

$$\mathcal{E}^2 = \iint (\alpha^2 \mathcal{E}_c^2 + \mathcal{E}_b^2)\, dxdy \tag{22}$$

The minimization is to be accomplished by finding suitable values for the optical flow velocity $(u, v)$. Using the calculus of variation we obtained

$$E_x u^2 + E_x E_y v = \alpha^2 u \nabla^2 - E_x E_t \tag{23}$$

$$E_x E_y u + E_y^2 v = \alpha^2 u \nabla^2 - E_y E_t$$

Using the approximation to the Laplacian introduced in the previous section and then solving for $u$ and $v$ we found that

$$(\alpha^2 + E_x^2 + E_Y^2)u = +(\alpha^2 + E_y^2)\bar{u} - E_x E_Y \bar{v} - E_x E_t \tag{24}$$

$$(\alpha^2 + E_x^2 + E_Y^2)\upsilon = +(\alpha^2 + E_X^2)\bar{\upsilon} - E_x E_Y \bar{u} - E_x E_t$$

Difference of Flow at a Point from Local Average

$$(\alpha^2 + E_x^2)u + E_x E_Y \upsilon = (\alpha^2 \bar{u} - E_x E_t) \tag{25}$$

$$E_x E_Y u + (\alpha^2 + E_Y^2)\upsilon = (\alpha^2 \bar{\upsilon} - - E_x E_t)$$

These equations can be written in the alternate form

$$(\alpha^2 + E_x^2 + E_Y^2)(u - \bar{u}) = -E_X \{E_X \bar{u} + E_Y \bar{\upsilon} + E_t\} \tag{26}$$

$$(\alpha^2 + E_x^2 + E_Y^2)(\upsilon - \bar{\upsilon}) = -E_y \{E_X \bar{u} + E_Y \bar{\upsilon} + E_t\}$$

This shows that the value of the flow velocity $(u, v)$ which minimizes the error $\mathcal{E}^2$ lies in the direction towards the constraint line along a line that intersects the constraint line at right angles.

## 4. PROPOSED SYSTEM

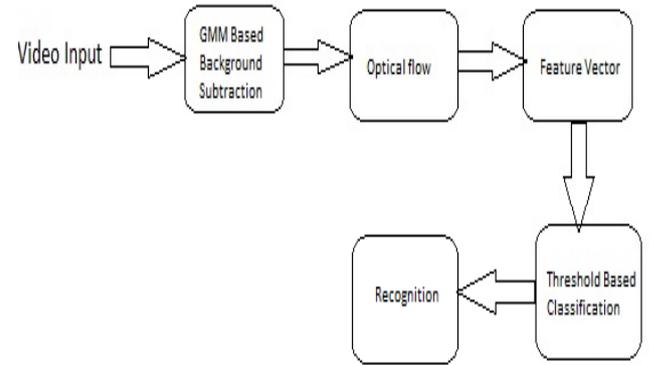

**Fig 2: Block diagram**

Our proposed system composed the general framework:

At first a frame (image) from the webcam is captured. It acted as the video input on which processing was done. The main information was collected from the image, so subtraction of the background for the entire process was done. There are a lot of techniques which provide helpful tool for image subtraction. Among those Gaussian Mixture Model (GMM) [27] gave the best result.

After using GMM it is tried to remove the background from the image. Some experimental results after background subtraction are shown

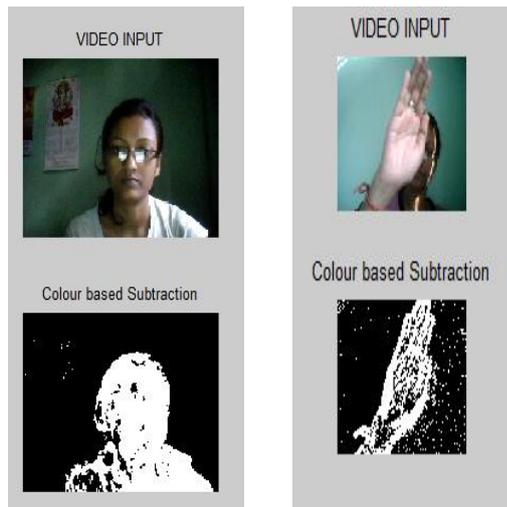

**Fig 3: Video input and background subtracted image**

After using Gaussian mixture model (GMM) we applied the Horn-Schunk optical flow algorithm to the foreground to determine the movement of the given input. Using optical flow algorithm we determined the movement between the frames of images of the video. An idea is given below regarding the intensity variation between the frames of inputed video-

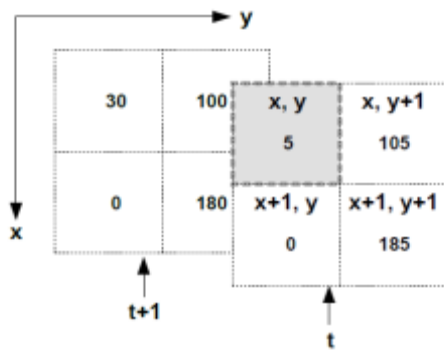

$I_X = ((105-50)+(185-0)+(100-30)+(180-0))/4 = 133.75$

$I_Y = ((0-5)+(185-105)+(0-30)+(180-100))/4 = 31.25$

$I_t = ((30-5)+(0-0)+(100-105)+(180-185))/4 = 3.75$

**Fig 4: The partial derivative of image brightness at point (x, y).**

This intensity variation is used to determine the optical flow of the foreground.

A lot of features are accompanied with the image. The information from the foreground is used to determine the movement of the head and that head movement ultimately provides us the corresponding gesture. Here the optical flow vectors acted as the feature vector.

Experimentally set threshold is used to identify the gesture from the summed optical flow vectors. In our experiment we tried to determine the movement of head to left, right, up and down. This movement in turn lead us to determine different kind of gesture such as yes, no, come etc. The optical flow we are implementing in our project is basically working on binary images. Further improvement will be done in nearby future. It is hoped that it will provide us greater accuracy.

## 5. RESULT AND DISCUSSION

After implementation of background subtraction and optical flow, we calculated the optical flow vector. In Table 1 sx and sy are the sum of the both optical flow vectors. Some summed values obtained shown below:

**Table 1. Head movement and corresponding values of sx and sy**

| Movements | Value of sx | Value of sy | Result | Recognition |
|---|---|---|---|---|
| | 0.0018 | 0.6535 | right | |
| | 0.0022 | 0.6550 | right | |
| | 0.0027 | 0.6564 | right | RIGHT |
| | 0.0031 | -0.0630 | down | |
| | -0.2038 | -0.2887 | left | |
| | -0.0495 | -0.1421 | left | |
| | -0.0080 | -0.1397 | left | LEFT |
| | 0.0062 | -0.1075 | left | |
| | 0.3968 | -0.0634 | down | |
| | 0.4085 | 0.0627 | right | |
| | 0.4602 | -0.0621 | down | DOWN |
| | -0.4627 | -0.0630 | down | |
| | -0.0015 | 0.1430 | up | |
| | -0.0035 | 0.1442 | up | |
| | -0.0046 | 0.1446 | up | UP |
| | -0.0072 | 0.1502 | up | |

From the above given value it can be concluded that-

**Table 2. Results from sx and sy**

| Value of sx | Value of sy | Conclusion |
|---|---|---|
| + ve | -ve | Down |
| -ve | -ve | Left |
| +ve | +ve | Right |
| -ve | +ve | Up |

The result can be shown as below

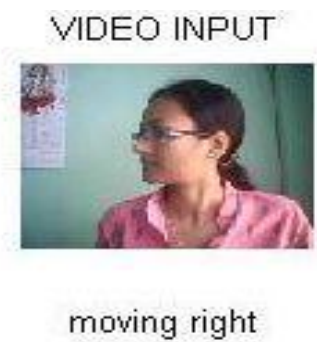

**Fig 5: Head is moving toward right**

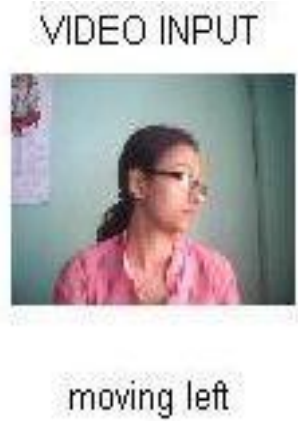

**Fig 6: Head is moving toward left**

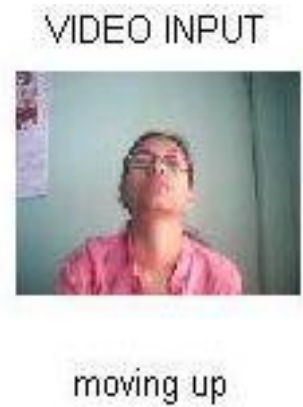

**Fig 7: Head is moving toward up**

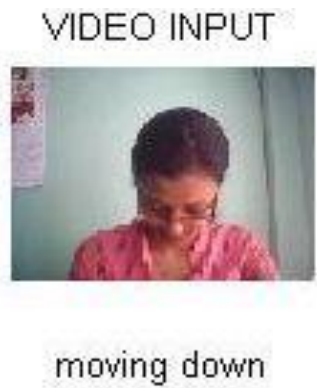

**Fig 8: Head is moving toward down**

Now from the movement of the head some sort of gestures can be determined. Such as up-down movement, left- right movement can be compared with yes and no gesture respectively.

**Table 3. Gesture detection**

| Head movement | Corresponding gesture |
|---|---|
| Left to right | NO |
| Up down | YES |

The success rate has been calculated and found to be 92.5% with the use of threshold based classification. The time taken to process was obtained to be 8.04 frames per second.

## 6. CONCLUSION AND FUTURE WORK

The system has been executed using MATLAB version 7.6.0.324(R2008a) and Intel(R) Core™2 Duo CPU @2.20 GHz processor machine, Windows 7 Ultimate (32 Bit),2.00 RAM and 0.3M Integrated Camera. In this system we have only considered the gesture form the video or moving scene. This system can be further upgraded to give order and control robots. Some more applications are that this Head Gesture Recognition system can be used in video games. Instead of using the mouse or keyboard, we can use some pre-defined gesture to play any game. Also, this system can be used to operate any electronic system by just showing gestures. Another application is that this can be used for security and authorization by keeping any particular gesture as the password. Replacement of mouse, gaming and entertainment industry are other areas in which gesture application is already in use. By improving the success rate, making background subtraction more proper, this work can be done more versatile. In nearby future optical flow shall be implemented on gray images. The success rate and the comparison of this project with some previous work are given below:

**Table 4. Success rate of classification**

| Experiments | Number of videos on which experiment was done | Number of correct recognition | Success rate |
|---|---|---|---|
| Left | 10 | 10 | 100% |
| Right | 10 | 8 | 80% |
| Up | 10 | 10 | 100% |
| Down | 10 | 9 | 90% |

**Table 5. Comparative study between our work and other approaches**

| Name of the technique used | Success rate | Comparative study |
|---|---|---|
| Optical flow [3] | 97.4% | Noise caused unintended head motion |
| PCA[11] | 94.4% | Error in reconstruction increases when eigentrajectories deceases |
| Lucas kanade's algorithm[17] | 100% | Fails to compute high displacement rate |
| SVM[21] | 85.3% | High false positive rate |
| Our work | 92.5% | • Less noise<br>• Improved processing rate<br>• Accurate time derivatives |

## 7. ACKNOWLEGDEMENT
We gratefully acknowledge the support provided by Dr Sunandan Baruah, HOD, ECE, Don Bosco College of Engineering and Technology. Also, we would like to acknowledge the reviewers for their valuable comments and suggestions that helped to improve the quality of the manuscript.